\documentclass[12pt]{iopart}

\usepackage{graphicx}
\usepackage{dcolumn}
\usepackage{bm}

\include{epsf} 

\begin{document}

\title{Lexical evolution rates by automated stability measure}
\author{Filippo Petroni}  
\address{DIMADEFA, Facolt\`a di Economia, 
Universit\`a di Roma "La Sapienza", 
I-00161 Roma, Italy}
\author{Maurizio Serva}
\address{Dipartimento di Matematica,
Universit\`a dell'Aquila,
I-67010 L'Aquila, Italy}
\bigskip

\date{\today}

\begin{abstract} 
Phylogenetic trees can be reconstructed from the matrix 
which contains the distances between all pairs of languages 
in a family.   Recently, we proposed a new method which 
uses normalized Levenshtein distances among words with 
same meaning and averages on all the items of a given list. 
Decisions about the number of items in the input 
lists for language comparison have been debated since 
the beginning of glottochronology.
The point is that words associated to
some of the meanings have a rapid lexical evolution.
Therefore, a large vocabulary comparison
is only apparently more accurate then a smaller one
since many of the words do not carry any useful information.
In principle, one should find the optimal length
of the input lists studying the stability
of the different items.
In this paper we tackle the problem 
with an automated methodology only based 
on our normalized Levenshtein distance.
With this approach, the program of an automated
reconstruction of languages relationships is completed.
\end{abstract}

\maketitle
 
\section{Introduction}
Glottochronology tries to estimate the time at which languages 
diverged with the implicit assumption that vocabularies change 
at a constant average rate.
The concept seems to have his roots in the work of 
the French explorer Dumont D'Urville.
He collected comparative words lists of various
languages during his voyages around the Astrolabe 
from 1826 to 1829 and, in his work about 
the geographical division of the Pacific \cite{Urv},
he introduced the concept of lexical cognates and
proposed a method to measure the degree of relation 
among languages.
He used a core vocabulary of 115 basic terms
which, impressively, contains all but three 
the terms of the Swadesh 100-item list.
Then, he assigned a distance from 0 to 1 to any pair 
of words with same meaning 
and finally he was able to resolve the
relationship for any pair of languages.
His conclusion is famous: 
{\it La langue est partout la m\^eme}.

The method used by modern glottochronology,
developed by Morris Swadesh \cite{Sw} in the 1950s,
measures distances from the percentage 
of shared  $cognates$. 
Recent examples are the studies of Gray and Atkinson 
\cite{GA} and Gray and Jordan \cite{GJ}.
Cognates  are words inferred to have a common
historical origin,
and cognacy decisions are made by trained 
and experienced linguists. 
Nevertheless, the task of counting the number of 
cognate words in a list is far from being
trivial and results may vary for different studies.
Furthermore, these decisions may imply an enormous
working time. 
  
Recently, we proposed a new automated method 
\cite{SP, SP2} which has some advantages,
the first is that it avoids subjectivity
the second is that results can be replicated 
by other scholars assuming that
the database is the same,
the third is that no specific linguistic knowledge
is requested, and the last, but surely not 
the least, is that it allows for rapid comparison of
a very large number of languages.
We applied our method to the 
Indo-European and the Austronesian groups 
considering, in both cases, fifty different languages.
 
In our work, 
we defined the distance of two languages 
by considering a normalized Levenshtein
distance among words with the same meaning and we 
averaged on the two 
hundred words contained in a 200 words list 
\cite{footnote}.
The normalization, which
takes into account word length,
plays a crucial role,
and no sensible results would have been found without.

Almost at the same time, the above described
automated method was used and developed 
by another large group of scholars \cite{Bak, Hol}. 
In their work, they used lists of 40 words 
while we used lists of 200. 
Their choice was taken according to a careful 
study of the stability of different words.

Decisions about the number of words in the input 
lists for languages comparison was debated since 
the beginning of glottochronology, Swadesh himself
switched from 200 words lists to 100 words ones.
The point is that a large vocabulary comparison
is only apparently more accurate, on the contrary,
many of the words do not carry any information
on language similarity,
and their inclusion in the lists has the
only effect of increasing the error noise that
may hide the wanted results.
In fact, words evolve because of lexical changes, 
borrowings and replacement at a rate which
is not the same for all of them.
The speed of lexical evolution, is different for different
meanings and it is probably
related to the frequency of use
of the associated words\cite{PAM} .
Those meanings with a high rate of change turns to be
useless to establish relationships among languages.
Furthermore the study of words stability has an interest 
in itself since it may give strong information on the 
activities which are at the core of the 
behavior of a social or ethnic group.

The idea of inferring the stability of an item from its similarity 
in related languages goes back a long way in the lexicostatistical 
literature\cite{K,O,TD}.
In this paper we tackle this problem 
with an automated methodology based
on normalized Levenshtein distance.
For any meaning, and any linguistic group,
we are able to find a number which measure
its stability (or degree of evolution speed)
in a completely objective and
reproducible manner. With this approach,
the program of an automated reconstruction of 
languages relationships is completed.
This is different from the approach in
\cite{Bak, Hol} since they have a combined approach,
their lists are chosen according to a stability study
which makes use of cognates, and then they 
reconstruct the languages phylogeny
by using Levenshtein distance.

In the next section we define the lexical distance between words and
we also sketch our method for computing the time divergence between 
languages. 
Section 3 is the core of the paper,
there we define the automated stability measures
of the meanings and we make some preliminary study
concerning distribution and ranking
of stability for Indo-European languages. 
In section 4 we study correlations and 
Fouldy-Robinson differences associated to
lists of different length.
We take here the decision about the meanings 
that should be included in the lists.
Conclusions and outlook are in section 5.

\section{Definition of distance}
We define here the lexical distance between two 
words which is a variant of the Levenshtein (or edit) distance. 
The Levenshtein distance is simply the minimum
number of  insertions, deletions, or substitutions of a
single character needed to transform one word into the other.
Our definition is taken as the edit distance divided 
by the number of characters in the longer of the two
compared words.

More precisely, given two words $\alpha_i$ and $\beta_j$
their distance $D(\alpha_i, \beta_j)$ is given by

\begin{equation}
D(\alpha_i, \beta_j)= 
\frac{D_l(\alpha_i, \beta_j)}{L(\alpha_i, \beta_j)}
\label{wd}
\end{equation}
where $D_l(\alpha_i, \beta_j)$ is the 
Levenshtein distance between the two words
and $L(\alpha_i, \beta_j)$ is the
number of characters of the longer of the two words
$\alpha_i$ and $\beta_j$.
Therefore, the distance can take any value between 0
and 1. Obviously  $D(\alpha_i, \alpha_i)=0$ .

The normalization is an important novelty and it
plays a crucial role;
no sensible results can been found without\cite{SP, SP2}.

We use distance between pairs of words, as defined above, 
to construct the lexical distances of languages. 
For any pair of languages, the first step is to compute
the distance between  words corresponding to the same meaning 
in the Swadesh list. 
Then, the lexical distance between each
languages pair is  defined as the average of the 
distance between all words\cite{SP, SP2}.  
As a result we have a number between 0 and 1 which 
we claim to be the lexical distance between two languages. 
 
Assume that the number of languages is $N$
and the list of words for any language contains
$M$ items. 
Any language in the group is labeled a Greek letter
(say $\alpha$) and any word of that language by 
$\alpha_i$ with $1 \leq i \leq M$. 
Then, two words $\alpha_i$ and $\beta_j$  
in the languages $\alpha$ and $\beta$ 
have the same meaning 
(they corresponds to the same meaning) if $i=j$.

Then the distance between two languages is

\begin{equation}
D(\alpha, \beta)=  \frac{1}{M} \sum_i
D(\alpha_i, \beta_i)
\label{ld}
\end{equation}
where the sum goes from 1 to $M$.
Notice that only pairs of words with same meaning are 
used in this definition. This number
is in the interval [0,1], obviously $D(\alpha, \alpha)=0$.

The results of the analysis is a   
$N \times N $ upper triangular matrix
whose entries are the $N(N-1)$ non trivial lexical 
distances $D(\alpha, \beta)$ between all pairs in a group.
Indeed, our method for computing distances is a very 
simple operation, 
that does not need any specific linguistic knowledge 
and requires a minimum of computing time.

A phylogenetic tree can be constructed from 
the matrix of lexical distances $D(\alpha, \beta)$,
but this gives only the topology of the tree
whereas the absolute time scale is missing.
Therefore, we perform \cite{SP, SP2} a logarithmic transformation 
of lexical distances which is the analogous of
the adjusted fundamental formula of glottochronology\cite{ST}.
In this way we obtain a new
$N \times  N$ upper triangular matrix whose entries
are the divergence times
between all pairs of languages.  
This matrix preserves the topology of the lexical 
distance matrix but it also contains the information 
concerning absolute time scales.
Then, the phylogenetic tree can be straightforwardly constructed.

In \cite{SP, SP2} we tested our method 
constructing the phylogenetic trees of the Indo-European group 
and of the Austronesian group.
In both cases we considered $N=50$ languages.
The database\cite{footnote}  that we used in
\cite{SP, SP2}  is 
composed by $M=200$ words for any language.. 
The main source for the database for the Indo-European group 
is the file prepared by Dyen et
al. in \cite{D}.  
For the Austronesian group we used as the main source the 
lists contained in the huge database in \cite{NZ}.

Criticism has been made to our proposal \cite{Nic}
on based on the fact that our reconstructed tree
presents some incongruence as for example
the early separation of Armenian which is
not grouped together with Greek (which, 
in our tree separate just after Armenian).
Nevertheless, the structure of the top of
the Indo-European tree is debated and 
no universally accepted conclusion exists. 

In our previous work we have adopted the 
historically motivated choice of 
200 words lists with the meanings proposed by Swadesh. 
Our aim, in this paper, is
to establish in a objective manner the proper length and 
the composition of the lists.
In order to reach this goal we need to 
separately study the stability of any meaning.

\section{Stability of meanings}
We take now decisions concerning stability of meanings. 
Our aim is to obtain an automated procedure, which avoids, 
also at this level, the use of cognates.
For this purpose, it is necessary
to obtain a measure of the typical distance
of all pairs of words corresponding to a given 
meaning in a language family.
The meaning is indicated by the label $i$
and $\alpha_i$ is the corresponding
word in the language $\alpha$.
Therefore, we define the stability as:

\begin{equation}
S(i) = 1- \frac{2}{N(N-1)} 
\sum_{\alpha > \beta}
D(\alpha_i, \beta_i)
\label{si}
\end{equation}
where the sum goes on all possible $N(N-1)/2$
possible language pairs $\alpha$, $\beta$  
in the family using the fact that $D(\alpha_i, \beta_i)=D(\beta_i,\alpha_i)$.
 
With this definition, $S(i)$ is inversely proportional
to the average of the distances $D(\alpha_i, \beta_i)$
and takes a value between 0 and 1.
The averaged distance is smaller
for those words
corresponding to meanings with a lower rate of 
lexical evolution since they tend to remain
more similar in two languages. 
Therefore, to a larger $S(i)$ corresponds
a grater stability.

We computed the $S(i)$ for the 200 meanings 
of 50 languages of the Indo-European family.
To have a first qualitative understanding 
of the distribution of the $S(i)$ we plot the associated histogram in Fig 1.
We can see that there is a fat tail on the
right of the histograms indicating
that there are some meaning with a quite large stability.
This tail is at very variance with a standard Gaussian behavior. 
The same result are obtained if we consider
the Austronesian family instead.
We remark that similar plots were 
computed in \cite{PAM} were the rates of lexical evolution
are obtained by the standard glottochronology approach. 

\begin{figure}
\centering
\includegraphics[height=7cm]{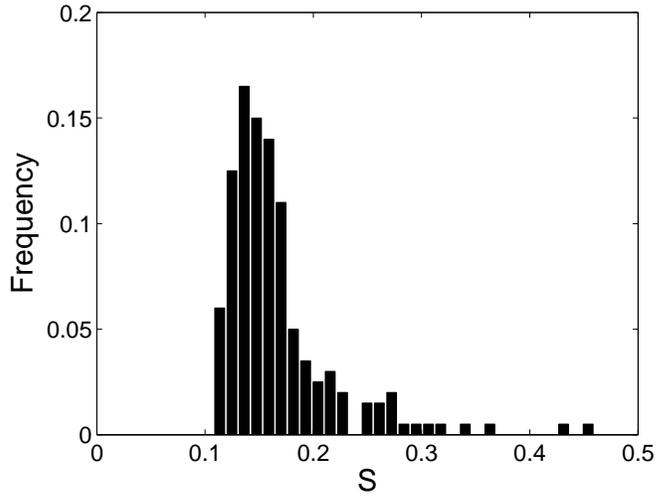}
\caption{Stability histogram of
meanings for Indo-European languages. 
The fat tail on the
right of the histogram indicates
that some items have a very large stability.}
\label{f1}
\end{figure}

To understand better the
behavior of the stability distribution, 
we plotted $S(i)$, in decreasing rank,
for the 200 meaning in the list.
In Fig. 2 are reported the data
concerning Indo-European family.
At the beginning the stability drops rapidly,
then, between the 50th position and the 180th
it decreases slowly and almost linearly
with rank, finally at the end the stability drops again.
We stress again that this behaviour is not Gaussian for which high and low stability part of the curve would be symmetric.
The curve is fitted by a straight line in the
central part of the data, between position
51 and position 180, in order to 
highlight the initial and final deviation
from the linear behavior.
We remark that the qualitative behavior for the Austronesian family
is exactly the same.

A preliminary conclusion is that one should surely keep all the 
meanings with higher information, take at least some of the 
most stable meanings in the linear part of the curve and exclude completely 
those meanings with lower information. 
Nevertheless, at this stage it is difficult to say
how many items should be maintained, since this number could be any between 50 and 180.

\begin{figure}
\centering
\includegraphics[height=7cm]{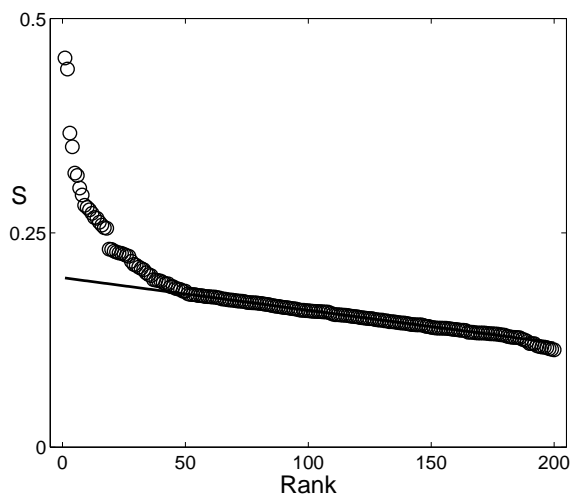}
\caption{Stability in a decreasing rank
for the 200 meanings of the Indo-European languages.
At the beginning stability has large values
but drops rapidly,
then, between the 50th position and the 180th
it decreases linearly, finally it drops again.
The straight line  between position
51 and position 180 underlines the initial and final deviation
from the linear behavior.}
\label{f2}
\end{figure}

It is necessary a deeper analysis of the 
stability to reach a conclusion.
Indeed, we need to know what is the minimum number of
meaning which allows for a precise computation 
of distances between languages
and, consequently, permits an accurate construction
of the phylogenetic tree.
In order to reach this goal we need a careful analysis of correlations 
among distances computed with the whole list
and distances computed with shorter lists.
It is also necessary to compare the phylogenetic
trees by a proper measure, the most natural being
the Robinson-Foulds difference.

\section{Correlations}
As mentioned in the previous section,
first of all we need to evaluate the impact of shorter lists on our 
estimate of the distances between languages. 
In order to reach this goal, 
we compute the correlation coefficient $c(n)$ between distances
$D(\alpha,\beta)$  obtained by
the whole list of 200 items and the distances 
$D_n(\alpha,\beta)$ obtained only by
the most $n$ stable  items (obviously,
$D(\alpha,\beta)$=$D_{200}(\alpha,\beta)$). 

The correlation coefficient $c(n)$ is computed in a standard way,
using averages over all possible
pairs of languages and it takes the value 1 only when there is
complete coincidence between $D_n(\alpha,\beta)$
and $D(\alpha,\beta)$ .
The correlation is plotted in Fig 3 for the Indo-European
family. Also in this case similar
results are found if the Austronesian family is considered.

From the figure one can observe that 
the correlation reaches a value larger than
$99\%$ with 100 meanings. 

\begin{figure}
\centering
\includegraphics[height=7cm]{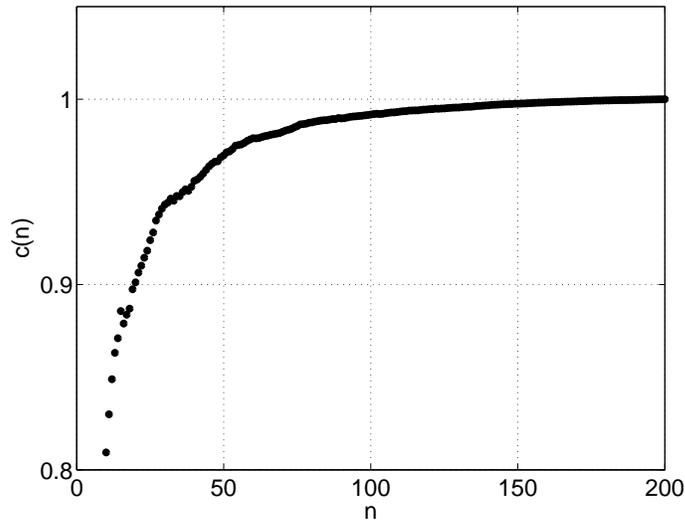}
\caption{Correlation coefficient
$c(n)$ of distances for Indo-European languages.
The coefficient $c(n)$ measures
the correlation between the distances estimated with
$n$ items and the distances estimated with 200 items.
$c(n)$ reaches a value larger than $99\%$ at $n = 100$.}
\label{f3}
\end{figure} 

The problem, is again that our choice for the length
of the lists depends on our choice for
the minimum excepted correlation coefficient.
If we accept $97\%$ we are satisfied by lists of 50 meanings
while if we need $99\%$ we have to take lists of 100 meanings.

To resolve this problem we
estimated the Robinson-Foulds difference\cite{RF}
between the trees generated stating from $D_n(\alpha,\beta)$
and the tree generated 
starting from $D(\alpha,\beta)$. 
The RF difference, which is plotted in Figure 3
for the Indo-European family,
measures the degree of similarity
between two trees. At lower values 
correspond trees which are more similar. 

As one can see from Fig. 4, the RF difference
drops rapidly until $n \sim 100$.
Than it remains almost constant
for all values greater then $n=100$
(the RF difference is equal to zero when $n=200$
but this is expected since $D_{200}(\alpha,\beta) =D(\alpha,\beta)$)
This result says  
that with 100 meanings one is able to capture all the 
information regarding languages distance 
and larger lists produce the same output.
In other words, the 100 meanings
which have been eliminated
carry small, if not vanishing, information.

\begin{figure}
\centering
\includegraphics[height=7cm]{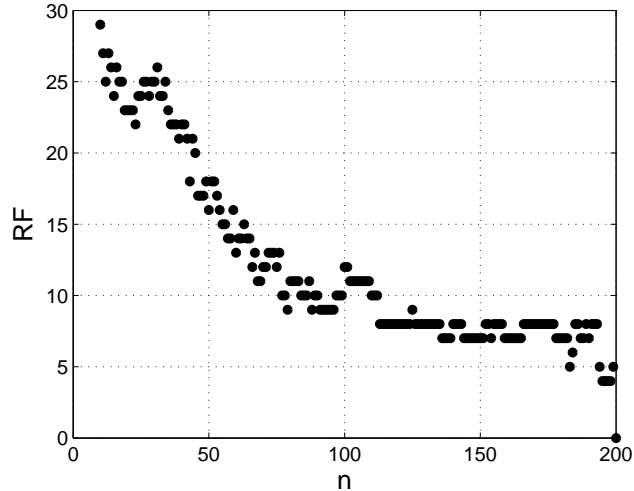}
\caption{Robinson-Foulds difference between trees
of Indo-European languages computed with lists of 200
items and lists of $n$ items.
The RF difference measures the degree of similarity
between trees. More similar trees have
a smaller difference. 
The RF difference
drops rapidly until $n \sim 100$,
than it remains almost constant
for all greater values of  $n$.}
\label{f4}
\end{figure}

The complete list of the most 100 stable terms 
for the Indo-European group can be found in Table 1.
The list is ordered by ranking, and the stability value
is written correspondingly to any item.
 
\begin{table} \caption{List of the 100 most stable meanings according to the $S(i)$ measure described in the text.}
\begin{center} 
\begin{tabular}{|l|l|l|l|l|l|l|l|} \hline Word & $S(i)$ & Word & $S(i)$ &Word & $S(i)$& Word & $S(i)$\\ \hline
you&0.45395&three&0.44102&mother&0.36627&not&0.35033\\new&0.31961&
nose&0.3169&four&0.30226&night&0.29403\\two&0.28214&name&0.27962&
tooth&0.27677&star&0.27269\\salt&0.26792&day&0.26695&grass&0.26231&
sea&0.25906\\die&0.25602&sun&0.25535&one&0.23093&feather&0.23055\\
give&0.22864&sit&0.22757&stand&0.22644&meat&0.2261\\long&0.22491&
five&0.22353&hand&0.22261&short&0.21676\\father&0.21319&smoke&0.21213&
far&0.20998&worm&0.20846\\dry&0.207&scratch&0.20343&person&0.20129&
when&0.20011\\wind&0.19535&snake&0.19485&sing&0.19434&stone&0.19369\\
suck&0.19196&mouth&0.19067&dig&0.19052&live&0.18716\\root&0.18715&
hair&0.18522&smooth&0.18457&water&0.18378\\tongue&0.18194&animal&0.1819&
year&0.17892&red&0.17815\\man&0.17801&tie&0.17789&snow&0.17697&
sew&0.17686\\there&0.17657&breathe&0.17578&flower&0.17566&mountain&0.17545\\
fruit&0.17508&bark&0.17502&sand&0.17443&leaf&0.1739\\warm&0.17283&
green&0.17269&liver&0.17205&hunt&0.17168\\sky&0.17156&know&0.17117&
bone&0.17056&spit&0.17036\\heart&0.17023&pull&0.16984&right&0.1689&
we&0.16858\\husband&0.16853&foot&0.1683&drink&0.16828&see&0.16764\\
lie&0.16763&fish&0.16693&woman&0.16656&louse&0.16624\\straight&0.16534&
yellow&0.16487&sleep&0.1643&black&0.16408\\who&0.16351&seed&0.16299&
wing&0.16288&cut&0.16245\\count&0.16173&thin&0.16156&sharp&0.1611&
float&0.16028\\fall&0.15968&earth&0.15965&kill&0.15926&burn&0.15918\\
\hline \end{tabular}
\end{center} \label{table1} \end{table}

In conclusion, one has to consider lists with the 100
meaning with higher stability, compute the matrix of lexical distances,
transform in the matrix of divergence times
and, finally, construct the tree.
The elimination of the 100 items with lower stability
has the positive effect of reducing the working time necessary
for an accurate check of all items and,
therefore, reducing errors due to misspelling
or inaccurate transliterations.
Furthermore, shorter lists allow for comparison of languages whose available vocabulary
is small.

\section{Discussion and conclusions}
In previous works \cite{SP, SP2} we proposed an automated method for
evaluating the distance between languages. 
Here we propose a method that is also automatic and gives 
lists of the mosts table meanings.
The novelty is that combining \cite{SP, SP2} with 
the results presented here everything can be done
automatically. 
Stable meanings, distances, divergence times and phylogenetic 
trees can be all obtained by using simple objective arguments
based on normalized Levenshtein distance.

We do not claim that our combined method produces better 
results then the standard glottochronology approach,
but surely comparable.
The advantages of this approach can be summarized here as follows: it avoids subjectivity
since all results can be replicated by other scholars assuming that
the database is the same; it allows for rapid comparison of
a very large number of languages; can be used also for those languages groups for which the use of cognates is very complicated or even impossible. 
In fact, the only work is to prepare the lists, while all the remaining work is made by
a computer program.

We would like to mention that
recently, together with other scholars \cite{Bl},
we have applied the method described here
as a starting point for
a deeper analysis of relationships among
languages. The point is that
a tree is only an approximation,
which, obviously, skips more complex
phenomena such as horizontal transfer.
These phenomena are reflected into 
the matrix of distances as deviations from
the ultra-metric structure.
It seems that the approach in\cite{Bl} allows for some more 
accurate understanding
of some important topics, such as migration patterns 
and homelands locations of families of languages.

\section*{Acknowledgments}
We warmly thank S$\o$ren Wichmann for helpful discussion.
We also thank Philippe Blanchard, Armando Neves, Luce Prignano and
Dimitri Volchenkov for critical comments on many aspects of the paper.
We are indebted with S.J. Greenhill, R. Blust and R.D.Gray,
for the authorization to use their: 
{\it The Austronesian Basic Vocabulary Database},
http://language.psy.auckland.ac.nz/austronesian
which we consulted in January 2008.

\newpage

\end{document}